# Precision-Aware Video Compression for Reducing Bandwidth Requirements in Video Communication for Vehicle Detection-Based Applications

**Abyad Enan***
Glenn Department of Civil Engineering
Clemson University, Clemson, South Carolina, 29634
Email: aenan@clemson.edu

**Jon C Calhoun, Ph.D.**
Associate Professor
Holcombe Department of Electrical and Computer Engineering
Clemson University, Clemson, South Carolina, 29634
Email: jonccal@clemson.edu

**Mashrur Chowdhury, Ph.D.**
Professor
Glenn Department of Civil Engineering
Clemson University, Clemson, South Carolina, 29634
Email: mac@clemson.edu

Word Count: 6567 words + 0 table (250 words per table) = 6,567 words

*Corresponding Author

## ABSTRACT

Computer vision has become a popular tool in intelligent transportation systems (ITS), enabling various applications through roadside traffic cameras that capture video and transmit it in real time to computing devices within the same network. The efficiency of this video transmission largely depends on the available bandwidth of the communication system. However, limited bandwidth can lead to communication bottlenecks, hindering the real-time performance of ITS applications. To mitigate this issue, lossy video compression techniques can be used to reduce bandwidth requirements, at the cost of degrading video quality. This degradation can negatively impact the accuracy of applications that rely on real-time vehicle detection. Additionally, vehicle detection accuracy is influenced by environmental factors such as weather and lighting conditions, suggesting that compression levels should be dynamically adjusted in response to these variations. In this work, we utilize a framework called Precision-Aware Video Compression (PAVC), where a roadside video camera captures footage of vehicles on roadways, compresses videos, and then transmits them to a processing unit, running a vehicle detection algorithm for safety-critical applications, such as real-time collision risk assessment. The system dynamically adjusts the video compression level based on current weather and lighting conditions to maintain vehicle detection accuracy while minimizing bandwidth usage. Our results demonstrate that PAVC improves vehicle detection accuracy by up to 13% and reduces communication bandwidth requirements by up to 8.23× in areas with moderate bandwidth availability. Moreover, in locations with severely limited bandwidth, PAVC reduces bandwidth requirements by up to 72× while preserving vehicle detection performance.

**Keywords:** Bandwidth reduction, Lossy compression, Precision-aware video compression, Vehicle detection, Vision-based object detection.

# INTRODUCTION

In the United States, transportation mishaps claimed the lives of over 370,000 people over the last decade, between 2011 and 2020, with roadway-related crashes accounting for 94% of these deaths (*1*). In 2022, motor vehicle crashes claimed the lives of 42,795 people, according to the National Highway Traffic Safety Administration (NHTSA) (*2*). The World Health Organization (WHO) projects that 1.19 million individuals worldwide lose their lives in traffic-related incidents each year (*3*). A collision warning system could have prevented many of these roadway crashes. With the advancement of modern computer vision technologies, computer vision techniques have gained popularity in various applications within intelligent transportation systems (ITS). These applications include traffic management (*4*), traffic surveillance (*5*), parking monitoring (*6*), pedestrian safety (*7*), crash prevention (*8*), basic safety message generation (*9*), and vehicle navigation (*10*). In video-based safety applications, a roadside video camera provides a 360° view of vehicles. The captured video is then sent to a computing node through a wireless medium, where vehicle detection and vehicle safety applications are executed. These applications leverage the detected vehicles' kinematics to generate Basic Safety Messages that are based on SAE 2735 (*11*). If the safety application predicts any potential conflict, it sends a real-time safety warning wirelessly to at-risk vehicles (*9*).

**Figure 1** illustrates a video-based potential safety application system for vehicles, where a roadside camera captures footage of a roadway. As a video-based safety application relies on processing image frames from video data, the captured video needs to be sent to a roadside video processing unit or to the cloud. The time it takes for the image frames of the video to transmit from the camera to the processing unit increases with the resolution of the video. For example, a single 1080p HD image requires 6× more storage than a 480p SD image. However, the reliability of a video-based safety application depends on the timely detection of vehicles within a low latency threshold. Because reliability depends on meeting the real-time latency requirements specified by the application, increasing latency reduces the likelihood of reliably detecting vehicles. Additionally, a growing number of connected devices and high-resolution cameras may compete for limited communication bandwidth, which could hinder the functionality of safety-critical applications like a video-based safety warning system. Therefore, to ensure road safety, the effectiveness of the video-based safety warning system hinges on the efficient transmission of video data from cameras to processing units. This paper presents a lossy data compression approach that reduces the amount of video data that needs to be stored and significantly lowers the data transmission latency within the communication network between the camera and the roadside video processing unit of the transportation infrastructure.

Video compression trades computational time to reduce data size. Most video compression algorithms are lossy because they do not perfectly preserve the video's images. Trading inaccuracies in the image frames enables significant reductions in video size (*12*). However, as the level of loss increases, the quality of the video decreases. The video compression limit for a specific algorithm is defined by quantifying the allowable loss level. Metrics such as peak signal-to-noise ratio (PSNR), root mean square error (RMSE), and structural similarity index (SSIM) are frequently used to assess the degree of loss in video data (*13*). Lossy compression algorithms that guarantee a predetermined fixed level of loss in the compressed data are referred to as error-bounded lossy compression (EBLC) algorithms (*14*).

As video compression reduces the bandwidth necessary for its transmission by lowering video quality (*15*); however, as the video becomes more compressed and the quality degrades, it decreases object detection accuracy. Additionally, object detection accuracy in real-world outdoor conditions is further compromised by adverse weather and lighting (*7*, *9*). Therefore, to achieve reliable performance of the object detection model for safety applications, a single static error tolerance level for the compressed video's quality is insufficient. Adjusting the error tolerance according to the external environment ensures that the accuracy of vehicle detection does not decline under unfavorable conditions.

The objective of this study is to reduce bandwidth requirements in video transmission for vehicle detection-based applications in diverse weather and lighting conditions, in locations where available bandwidth is limited, by employing a dynamic compression method called precision-aware video compression (PAVC). In this method, the image frames of the video are compressed to varying quality levels dynamically based on the weather and lighting conditions, while ensuring that vehicle detection efficiency, mean average precision (mAP) remains high. To further improve this efficiency, the detection model is calibrated with compressed data at different quality levels for various weather and lighting conditions. The objective of our PAVC method is to maintain high mAP in vehicle detection across different weather and lighting conditions while reducing communication latency. The contributions of this study are as follows:

- Incorporate a precision-aware video compression (PAVC) technique for video transmission for video-based vehicle detection applications in a range of weather and lighting conditions for varying compression levels
- Produce a set of pretrained models for vehicle detection for various weather and lighting conditions and compression levels, meeting high mAP in vehicle detection
- Evaluate the performance of PAVC by conducting real-world field tests to determine whether it lowers the necessary transmission bandwidth for video communication

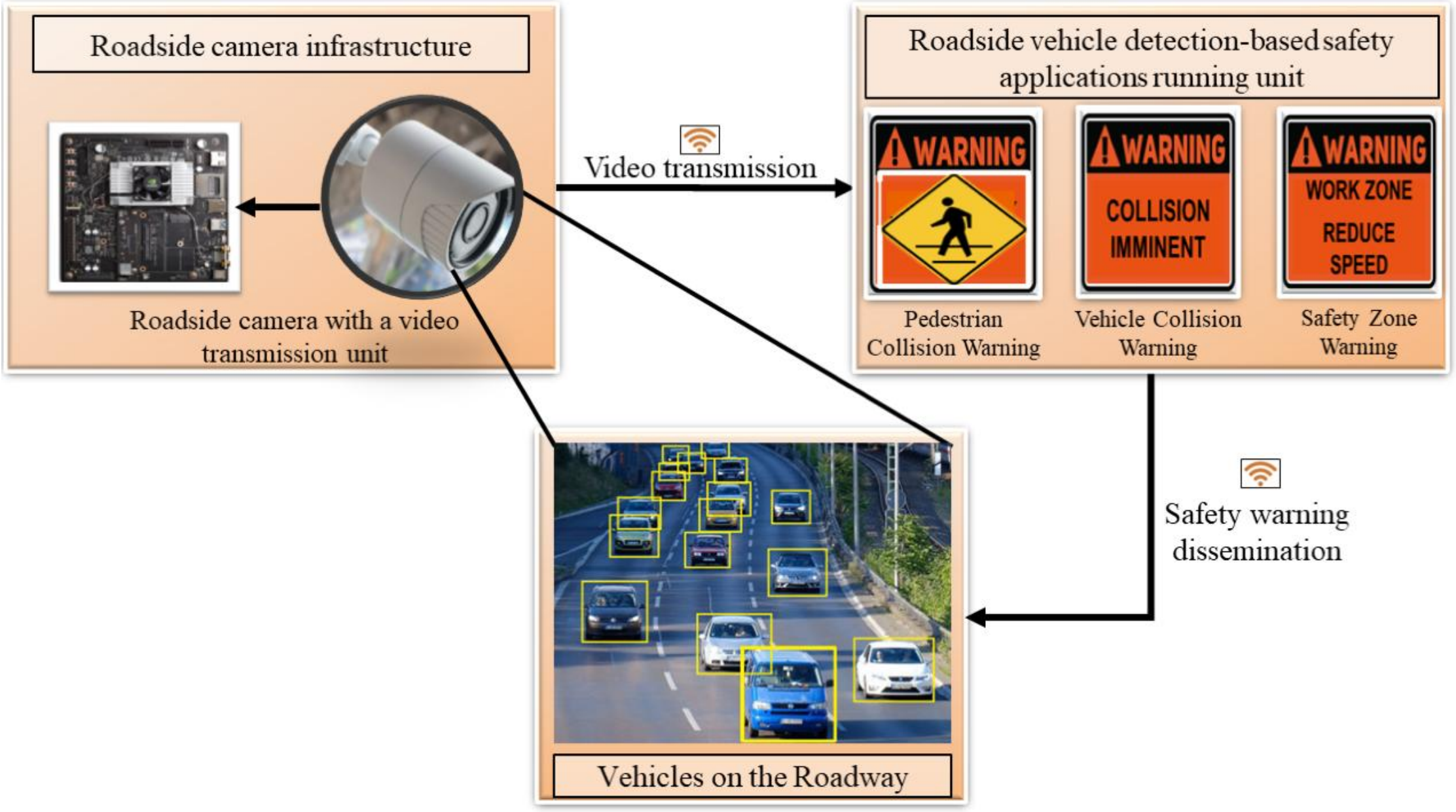

**Figure 1 A video-based vehicle safety application utilizing vehicle detection**

## RELATED WORK

Intelligent Transportation Systems (ITS) are designed to enhance transportation efficiency, safety, and surveillance by leveraging advanced technologies such as computer vision. Computer vision, a branch of artificial intelligence (AI), enables machines to interpret and process visual data in real-time, which is critical for ITS. In ITS, computer vision involves using cameras and algorithms to extract meaningful information from visual inputs. Applications include traffic monitoring (*5*), incident detection (*16*), smart parking management (*6*), automated tool collection (*17*), vehicle detection and classification (*17*), the perception module of autonomous vehicles (*10*), Basic Safety Message (BSM) generation (*9*), vehicle collision warnings (*8*), pedestrian safety (*7*), and more. Visual data is interpreted and analyzed using techniques such as machine learning, pattern recognition, and image processing.

In (*18*), the authors developed a traffic monitoring system, where congestion is detected, and traffic flow is monitored using computer vision. Vehicles are detected and counted using techniques such as motion analysis, object tracking, and background removal, which provide information on traffic density and flow patterns. Computer vision techniques are also incorporated for collision warnings, such as vehicle-to-vehicle collision warnings (*19*) and vehicle-to-pedestrian collision warnings (*7*). In (*7*), the authors developed a pedestrian safety warning system where a camera mounted at an intersection monitors the pedestrian crosswalk. As soon as a pedestrian is detected in the crosswalk by a pretrained YOLOv3 model, a safety warning is sent to vehicles approaching the intersection. The developed pedestrian safety alert system works under diverse environmental conditions. In (*9*), the authors developed a video-based Basic Safety Message (BSM) generation method for vehicles, where a video camera captures footage of vehicles and determines their kinematics, known as BSM. In this work, the authors used the YOLOv7 object detection method for vehicle detection under diverse weather and lighting conditions, which can be applied to potential safety applications or hazard prediction. In (*20*), the authors developed a vision-based incident detection system, which monitors illegal parking via video cameras. In this work, the authors used a deep learning-based novel voting method using an in-vehicle camera for illegal parking detection. An automated tool collection system developed by the authors in (*17*) incorporates computer vision techniques to collect tools automatically. Computer vision techniques are also used for traffic surveillance to detect distracted driving, aggressive maneuvers, or compliance with traffic signals, thereby contributing to road safety (*16*). Utilizing vision technologies, construction operations are monitored to ensure adherence to safety protocols and project schedules. These technologies can identify improper entry, misuse of equipment, and the use of safety gear (*21*).

Applications discussed in this section require camera data to be sent to a roadside unit or to the cloud through a communication medium to process videos. Sending high-quality video data requires significant communication bandwidth. Consequently, these applications may not be effective in areas with limited communication resources. This necessitates compression. Compression algorithms are classified into two categories based on their ability to preserve the fidelity of the data compressed. Lossless compressors such as Lempel-Ziv (LZ77) yield decompressed data that is bit-wise identical to the original (*22*). On the other hand, lossy algorithms have no such guarantee and result in inaccuracies in the decompressed data. Although lossy compression provides higher compression ratios and results in smaller data sizes, lossy compression (LC) increases the likelihood of inaccuracies in the decompressed data (*23*). Reducing video quality can eliminate important features in the video. Therefore, it is crucial to ensure that significant features are preserved during compression. For video compression, LC uses fewer bits to describe each frame, which introduces noise into the final product (*24*). Generally, a higher compression ratio corresponds to a greater loss in data fidelity (*25*). The most recent state-of-the-art LC algorithms, known as Error-Bounded Lossy Compression (EBLC) algorithms, allow users to control the amount of loss introduced during data compression (*26*). Modern video compression techniques, such as H.264 (*27*) and High-Efficiency Video Coding (HEVC) (*28*), are designed for high-resolution videos by packing more information into each compressed bit. Videos compressed with H.264 and HEVC achieve this by first identifying areas of intra-frame and inter-frame similarity. Then, transforms such as the discrete cosine transform (*29*) and coefficient encoding, or delta encoding, are applied to encode the changes between frames.

Previous works on lossy compression and object detection have explored methods to compress video quality while maintaining high object detection accuracy. One method in (*30*) employs object saliency maps as a preprocessing step to enhance video frame compression. This technique improves both transmission bitrate and object detection model accuracy. In contrast, another method in (*31*) found that temporal fluctuations in non-significant background segments of frames can degrade object detection efficiency. The authors propose an encoding technique to stabilize these temporal fluctuations, which enhances both bitrate and detection accuracy. Additionally, in (*14*), the authors integrated dynamic error-bounded lossy compression (DEBLC) to compress video while preserving high accuracy in pedestrian detection.

In our study, DEBLC from (*14*) is adopted rather than general compression to ensure that the video's quality does not affect the precision of vehicle detection. In this context, precision refers to the model's ability to produce bounding boxes that closely overlap with the ground truth, ensuring that detected vehicles are not only correctly classified but also accurately localized. High precision is crucial, as even small distortions in video quality can lead to inaccurate bounding boxes, reducing the detection reliability needed for downstream applications such as traffic monitoring or safety alerts (*9*). Our work builds upon the work in (*14*) to implement precision-aware video compression (PAVC) for vehicle detection-based applications, such as in (*9*). In (*14*), pedestrian data is collected from an intersection, which is then augmented to create images for different weather conditions for pedestrian detection model training. In their framework, a single image frame is transmitted first from the roadside camera to the roadside video compressor wirelessly, which then detects weather conditions and informs the roadside camera and video compressor of the appropriate compression level. The roadside compressor then compresses the video and sends it to the video processor. This process adds additional latency to the system. Additionally, the authors in (*14*), only theoretically demonstrated how DEBLC helps reduce communication latency without conducting field tests.

In our study, we overcome these limitations by collecting image data for vehicles from different roadways and angles to ensure that our model detects objects of various sizes and can precisely localize them. In our method, we perform the environmental and lighting condition detection in the roadside camera and compression unit, allowing it to compress the video immediately after capturing it based on the environmental conditions. This reduces the additional latency by eliminating the need to send a single image frame to the video processing unit and waiting for the compression ratio value in return. We also conduct field tests to evaluate the performance of our PAVC strategies and demonstrate bandwidth reduction.

## METHODS

In a video-based vehicle detection application, detection performance varies with different weather and lighting conditions. Degrading video quality leads to a decrease in vehicle detection precision. Therefore, it is crucial to dynamically compress the video to different levels based on varying weather and lighting conditions to maintain high precision in vehicle detection under all conditions.

**Figure 2** illustrates the overview of our PAVC application. A roadside video camera, equipped with video compression and weather detection capabilities, captures footage of the roadway. From the captured image frames, it detects the weather and lighting conditions. In this work, we have considered various weather and lighting conditions such as sunny, light-dark, medium-dark, heavy-dark, drizzle, moderate rain, and torrential rain. Based on the environmental condition, the video compressor then compresses the image frames of the video according to predetermined compression levels specific to the respective environmental condition to ensure that average precision in vehicle detection does not degrade below the target threshold, where the threshold is determined based on the performance of the object detector for uncompressed images to ensure that the detection performance does not fall due to video compression as presented later. In this study, the tolerance is based on the PSNR, which is the comparison between the raw video and the resulting compressed video. The PSNR value to which the video is compressed depends on the environmental conditions. H.264 is used for video compression; however, any other video compression method can be employed.

After compression, the compressed video is sent to a roadside computing node using wireless communication for vehicle detection. The roadside computing node has two components: (i) a vehicle detection model that detects vehicles from image frames of video and (ii) a set of pretrained models for vehicle detection under different environmental conditions calibrated for different compression levels. The PAVC method is independent of the vehicle detection model, meaning that any state-of-the-art (SOAT) vehicle detection method capable of processing video in real-time can be used. Based on the environmental condition of the received image frames, the respective pretrained model from the set of models is used to perform vehicle detection. Additionally, it determines the corresponding PSNR for the model that provides the largest bandwidth reduction while maintaining the same detection precision.

The video compressor integrated with the camera checks the environmental conditions in each frame and adjusts the video compression to the PSNR ratio at which vehicle detection precision remains constant while minimizing the video transmission bandwidth requirement through the wireless medium. This approach dynamically adjusts the compression level in response to environmental conditions and reduces the required communication bandwidth for video transmission. The subsequent subsections illustrate all the steps of our PAVC method.

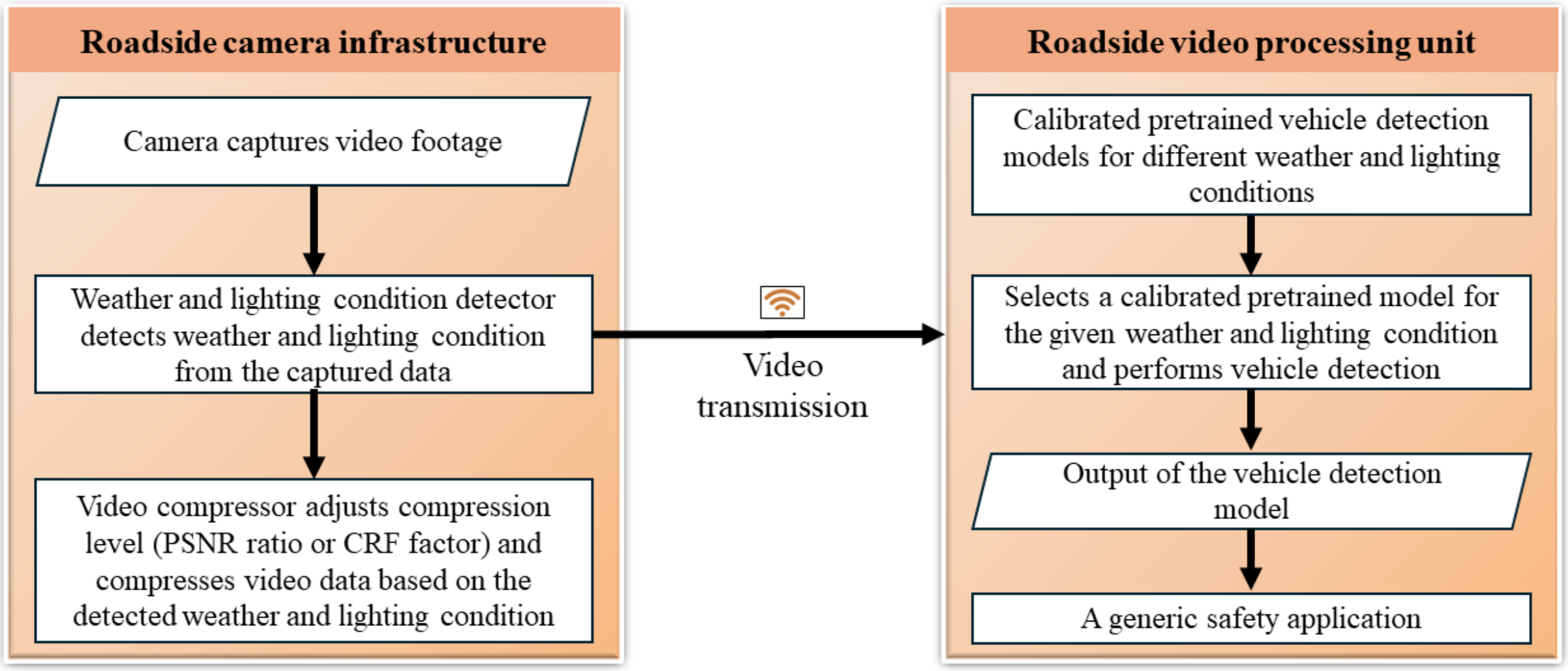

**Figure 2 Precision-aware video compression (PAVC) strategy**

**Weather and Lighting Condition Detector**

In this study, we use a Convolutional Neural Network (CNN) model for detecting environmental conditions. The classifier takes an image frame from the captured video as input and classifies the environmental condition into one of seven different weather categories: sunny, light-dark, medium-dark, heavy-dark, drizzle, moderate rain, and torrential rain. The size of the image frames that the model receives as input is 640×640×3 pixels (similar to the size of the images from the camera), and it outputs a 7×1 matrix. For example, if the output is $[0, 0, 0, 0, 1, 0, 0]^T$, it indicates that the weather is drizzling. We chose a simple CNN classifier for weather condition detection because it does not require significant computational resources.

**Vehicle Detector**

For vehicle detection, we use a state-of-the-art object detection method, YOLOv7 (You Only Look Once Version 7). YOLOv7 is an evolution of the YOLO family of models, which are well-regarded for their balance between speed and accuracy. YOLOv7 incorporates advancements from its predecessors (YOLOR, YOLOX, YOLOv3, Scaled-YOLOv4, YOLOv5, etc.) (*32*), making it one of the best choices for real-time video-based object detection. It is called "You Only Look Once" because the model processes image frames from a video in a single pass rather than examining an image multiple times or using a sliding window approach (*32*). By processing images only once, the computational overhead is significantly reduced, making it faster than non-YOLO methods. YOLOv7 has proven to perform admirably at a variety of frame rates. For example, YOLOv7 outperforms several modern models like YOLOR, YOLOX, Scaled-YOLOv4, YOLOv5, and DETR in both speed and precision among real-time object detectors operating at 30 frames per second (FPS) or higher on the GPU V100. It achieves an impressive accuracy of 56.8% Average Precision (AP) and a quick computational time of 28 ms (*32*). Due to its high frame rate, YOLOv7 is crucial for applications requiring real-time processing. YOLO models are designed to optimize the trade-off between speed and accuracy. While some non-YOLO methods might achieve higher accuracy in certain scenarios, they often do so at the expense of speed, making them less suitable for real-time applications. YOLO models can detect multiple objects in a single pass, and their architecture is well-suited for detecting objects of different scales and aspect ratios. Non-YOLO methods might require additional adjustments or separate processing to handle diverse objects effectively. Additionally, YOLOv7 is pretrained on the Microsoft COCO dataset with 80 classes of data. Retraining YOLOv7 is straightforward. YOLOv7 is developed using an innovative architecture, a compound model scaling strategy, and a uniquely implemented trainable bag-of-freebies method during training to increase detection accuracy without raising computational costs during inference (*32*). That is why we choose YOLOv7 over other object detection methods. However, our PAVC is independent of the object detection method. Any object detection model can be incorporated with our PAVC method.

### Precision-Aware Video Compression

The video data is compressed using FFmpeg in Python. The video compression level in Python is controlled with the CRF (Constant Rate Factor) value (*33*). A higher CRF value results in a lower PSNR value, meaning higher compression. The CRF value can range from 0 to 51, where 0 means no compression (no distortion) and 51 means maximum compression (most distortion).

To determine the optimal CRF for different environmental conditions, the CRF value is incremented, and the YOLOv7 model is retrained with compressed data, and then its mean average precision (mAP) is measured under different environmental conditions. If the mAP remains within the threshold accuracy, the CRF value is incremented again. The model is then retrained again with the newly compressed data from the similar training set, and its performance across all environmental conditions is evaluated again. This process continues for all environmental conditions as long as the mAP remains high and within the threshold accuracy.

This process is repeated for each weather and lighting condition until the performance goes below the threshold mAP, determined based on our analysis on uncompressed videos. If the performance for a given environmental condition degrades below the threshold mAP, the CRF value is not incremented further for that condition. In this case, the highest CRF value for which the model's mAP remains within the threshold is selected as the optimal CRF value for that environmental condition. This results in a total of seven CRF values (mentioned in the Analysis and Results section) being determined for the seven different weather and lighting conditions, and seven pretrained models are generated.

In our PAVC framework, when the roadside camera captures the video, the weather and lighting condition detector predicts the environmental condition. The video is then compressed using FFmpeg with the predetermined CRF value corresponding to the respective environmental conditions to ensure that vehicle detection performance does not fall below a threshold. The compressed video frames, along with an environmental condition variable (an integer ranging from 0 to 7 depending on the weather and lighting condition), are wirelessly transmitted, via Wi-Fi in our case, though any wireless method can be used depending on the available communication infrastructure, to the roadside video processing unit. This unit runs the YOLOv7 vehicle detection model, selecting the appropriate pretrained model tailored to the specific environmental condition. In this way, PAVC dynamically adjusts video compression based on environmental conditions to maintain reliable detection performance.

## ANALYSIS AND RESULTS

The subsequent subsections describe the data generation process, evaluation of the weather and lighting condition detector, calibration of the vehicle detection model with uncompressed data under diverse environmental conditions, vehicle detection model calibration for PAVC, and evaluation of PAVC for bandwidth reduction.

### Data Generation

In this study, images and video footage of vehicles on roadways, taken from roadside traffic cameras, are collected from Kaggle (*34*). These images and videos are captured under sunny weather conditions. The image dataset is already prepared for YOLO templates, so we do not need to manually draw the bounding boxes or label the dataset. There is a total of 341 images available, each containing multiple vehicles, with dimensions of 640×640×3.

To create images for different environmental conditions, such as light-dark, medium-dark, heavy-dark, drizzle, moderate rain, and torrential rain, we perform image augmentation to generate realistic images for diverse scenarios. We simulate nighttime darkness by adjusting the pixel values of the lightness channel in the image's HSL (hue, saturation, lightness) color space, thereby producing augmented data for model calibration (*35*). Various forms of rainy scenes are created by adding random small lines to the images and slightly blurring them to recreate a realistic rainy environment, depending on the intensity of the rain (*35*). Following the method described in Section IV, compressed image datasets with different compression levels (different CRF values) are created for all weather and lighting conditions. Similarly, image frames from the video data are augmented and compressed to create videos representing diverse environmental conditions for both uncompressed and compressed cases, which we use to evaluate the PAVC for required communication bandwidth reduction.

**Evaluation of Weather and Lighting Condition Detector**

The sunny weather images and the augmented images for diverse environmental conditions, totaling 341 × 7 = 2,387 images, constitute the dataset for the weather detector CNN model. This classification model categorizes images into one of the seven classes: sunny, light-dark, medium-dark, heavy-dark, drizzle, moderate rain, and torrential rain. The images are split into training, validation, and test sets at ratios of 75%, 15%, and 10%, respectively. The model is tested with the test dataset, and the results reveal that it classifies the weather conditions with 100% accuracy, showing no false positives or negatives, as depicted in the confusion matrix in **Figure 3**, where class label 0, 1, 2, 3, 4, 5, 6 represent sunny, light-dark, medium-dark, heavy-dark, drizzle, moderate rain, and torrential rain respectively.

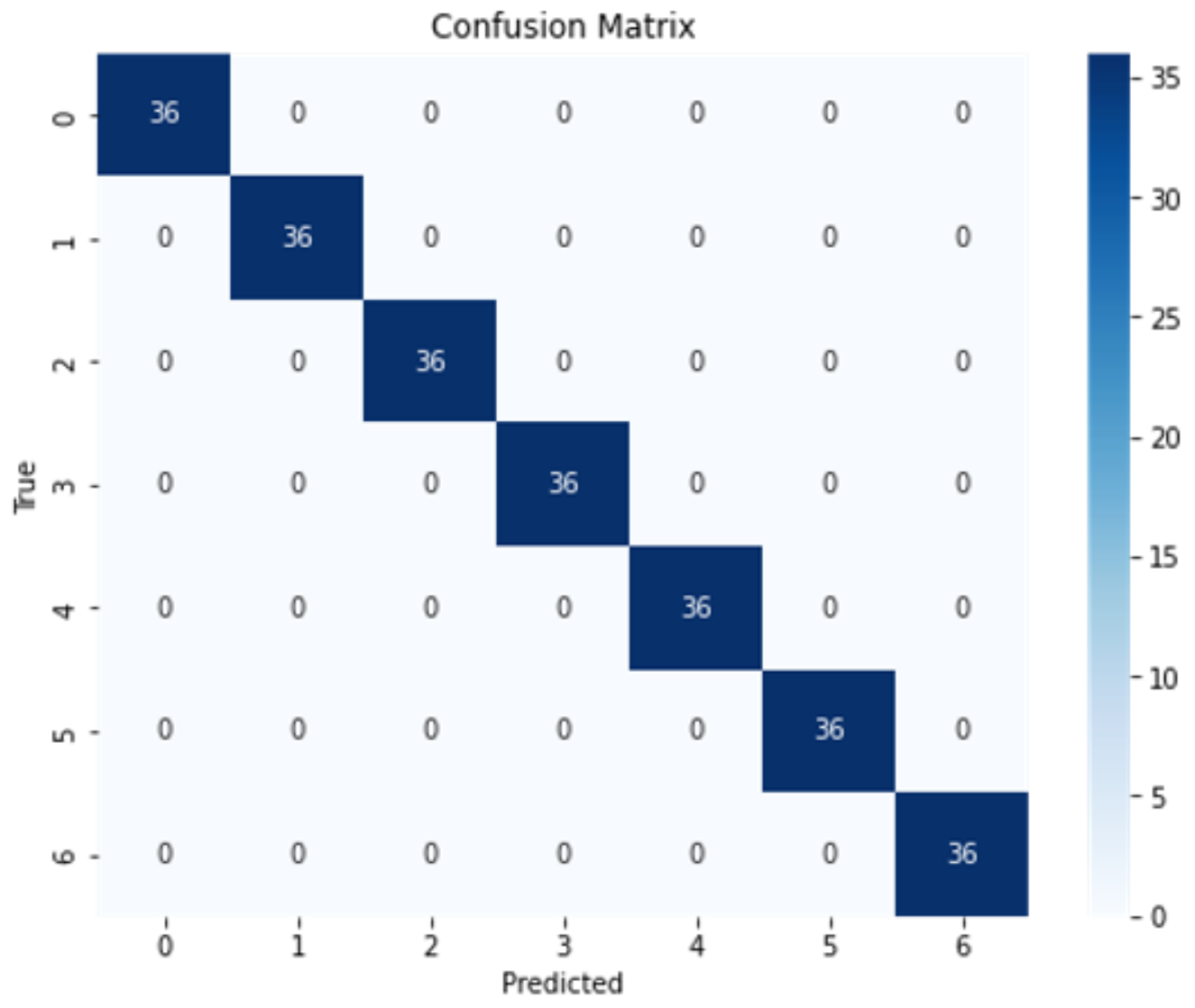

**Figure 3 Confusion matrix of the weather and lighting condition detector CNN model.**

**Vehicle Detection Model Calibration with Uncompressed Data Under Diverse Weather and Lighting Conditions**

We assess the performance of the YOLOv7 model for vehicle detection under different weather and lighting conditions by training it solely on sunny weather data to observe the effects of various environmental conditions on the vehicle detection performance. **Figure 4** shows the mAP of vehicle detection under different environmental conditions with no data compression. As the weather worsens, we observe a decrease in vehicle detection performance.

To increase the mAP, we retrain the YOLOv7 model with augmented, uncompressed data. **Figure 5** shows that mAP increases significantly after retraining the YOLOv7 model, up to a 13% increase for torrential rain. After retraining, the lowest mAP without compression is found under torrential rain conditions. This is chosen as the threshold for our application, which means that when the video is compressed, the compression level is adjusted to ensure that detection accuracy does not fall below the threshold of mAP 98.5%, maintaining consistent accuracy across all environmental conditions.

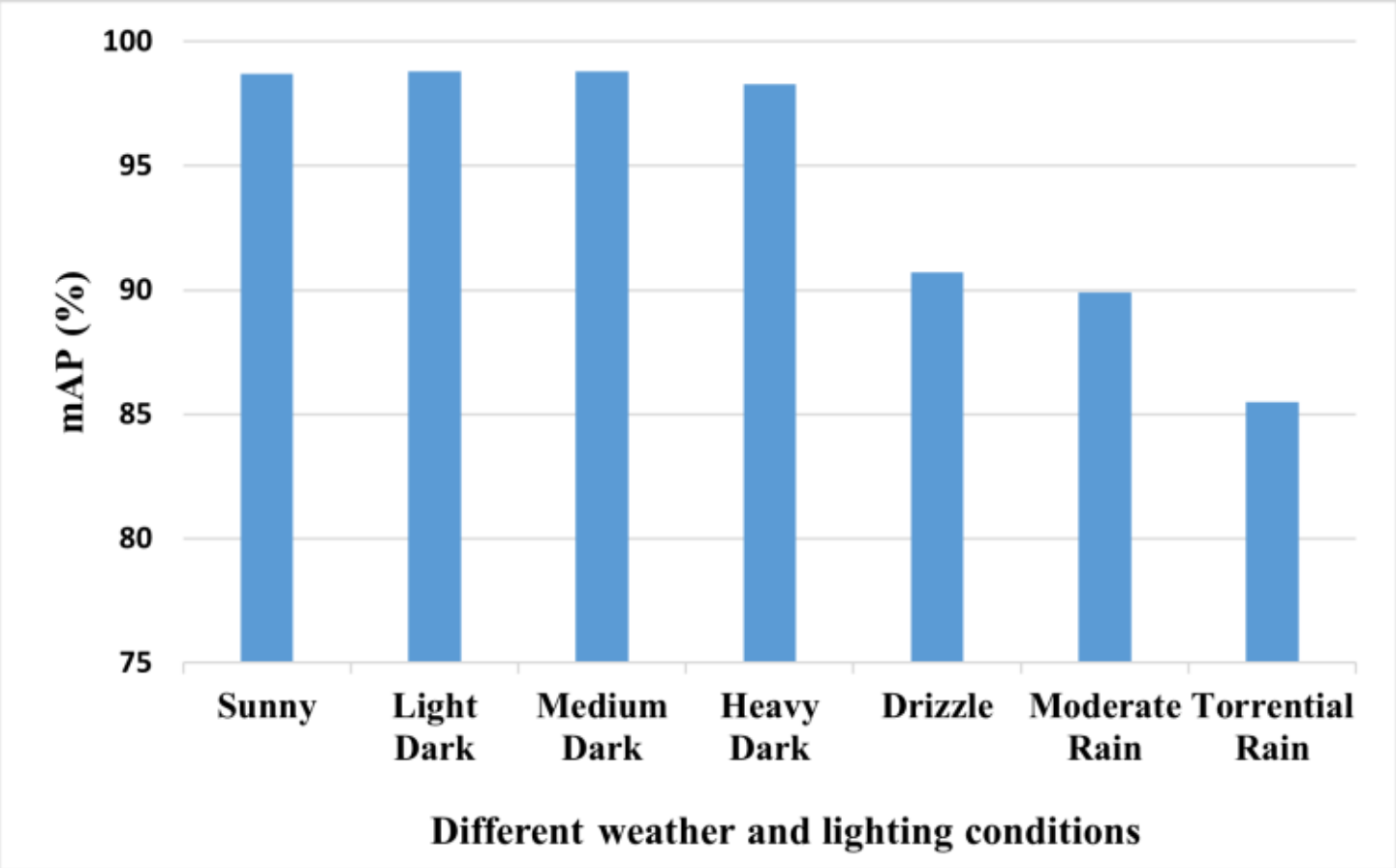

**Figure 4 Vehicle detection mAP for different weather and lighting conditions when using uncompressed images with a model trained solely on baseline data (sunny weather images).**

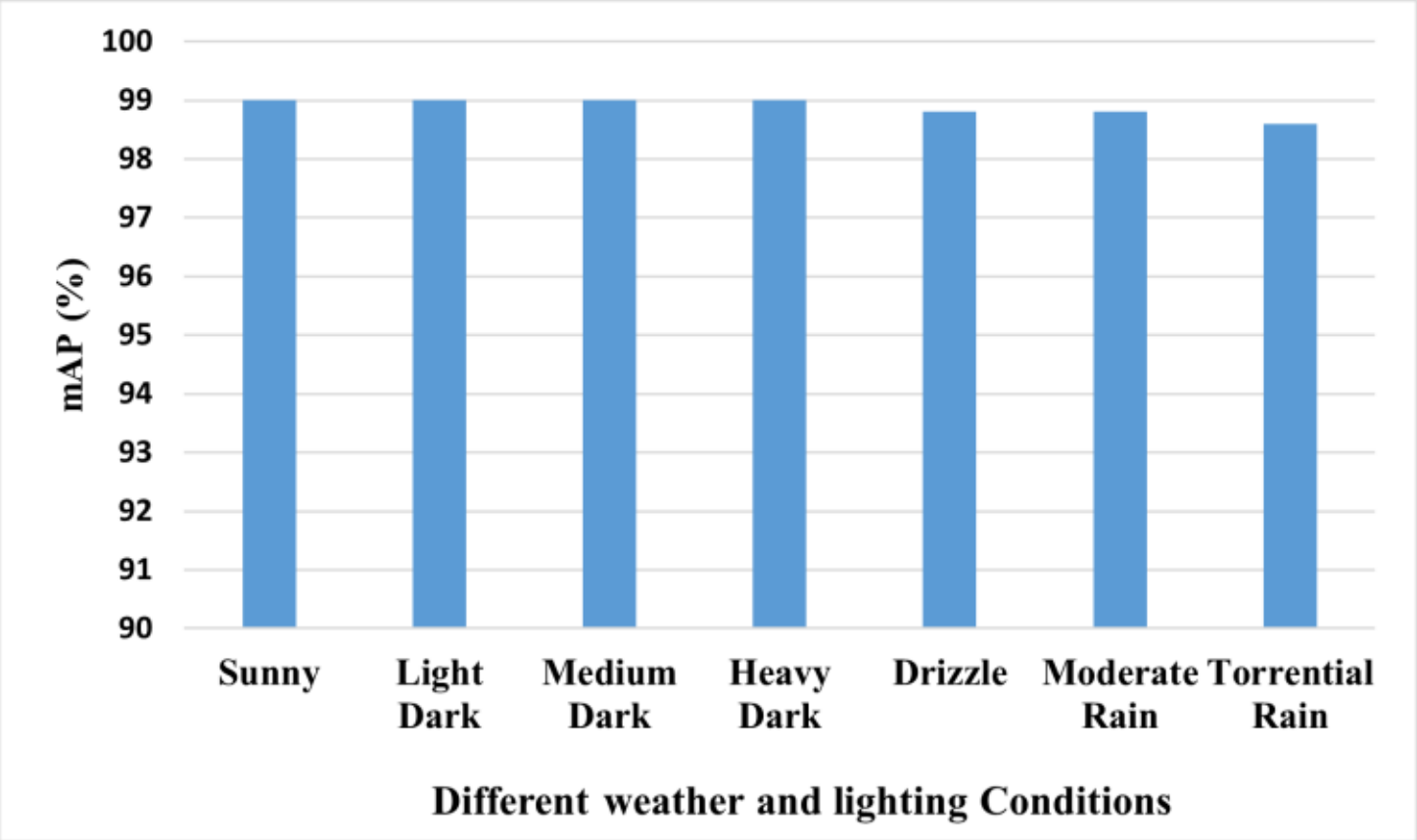

**Figure 5 Vehicle detection mAP for different weather and lighting conditions when using uncompressed images with a model trained with all-weather and lighting images.**

**Vehicle Detection Model Calibration for PAVC**

When compressing an image prior to vehicle detection in inclement weather, the accuracy of vehicle detection is further reduced. Consequently, the level of lossy compression and the environmental conditions affect the accuracy of the vehicle detection model. Therefore, it is crucial to train the vehicle detection model using data from various environmental scenarios and lossy compression levels.

The vehicle detection performance under different weather and lighting conditions at various compression levels (different CRF values) is evaluated. The steps mentioned in Section IV are followed to carry out the evaluation. The CRF value is incremented, and the model is retrained with the compressed data while tracking the mAP simultaneously. This process is repeated for all weather and lighting conditions to determine the highest level of compression (CRF value) for each condition at which the video can be compressed while keeping the performance of the model in detecting vehicles within the threshold.

**Figure 6** presents the detection accuracy after compressing the images of all weather and lighting conditions at different compression levels (CRF). The results reveal that vehicle detection accuracy drops below the threshold accuracy when compressing images under the following conditions:

- Sunny and light-dark weather: CRF > 40

- Medium-dark weather: CRF > 50
- Heavy-dark weather: CRF > 21
- Drizzle weather: CRF > 10
- Moderate rain: CRF > 7
- Torrential rain: CRF > 0

This implies that, in the PAVC method, the video should not be compressed more than CRF 40, 40, 50, 21, 10, 7, and 0 for sunny, light-dark, medium-dark, heavy-dark, drizzle, moderate rain, and torrential rain.

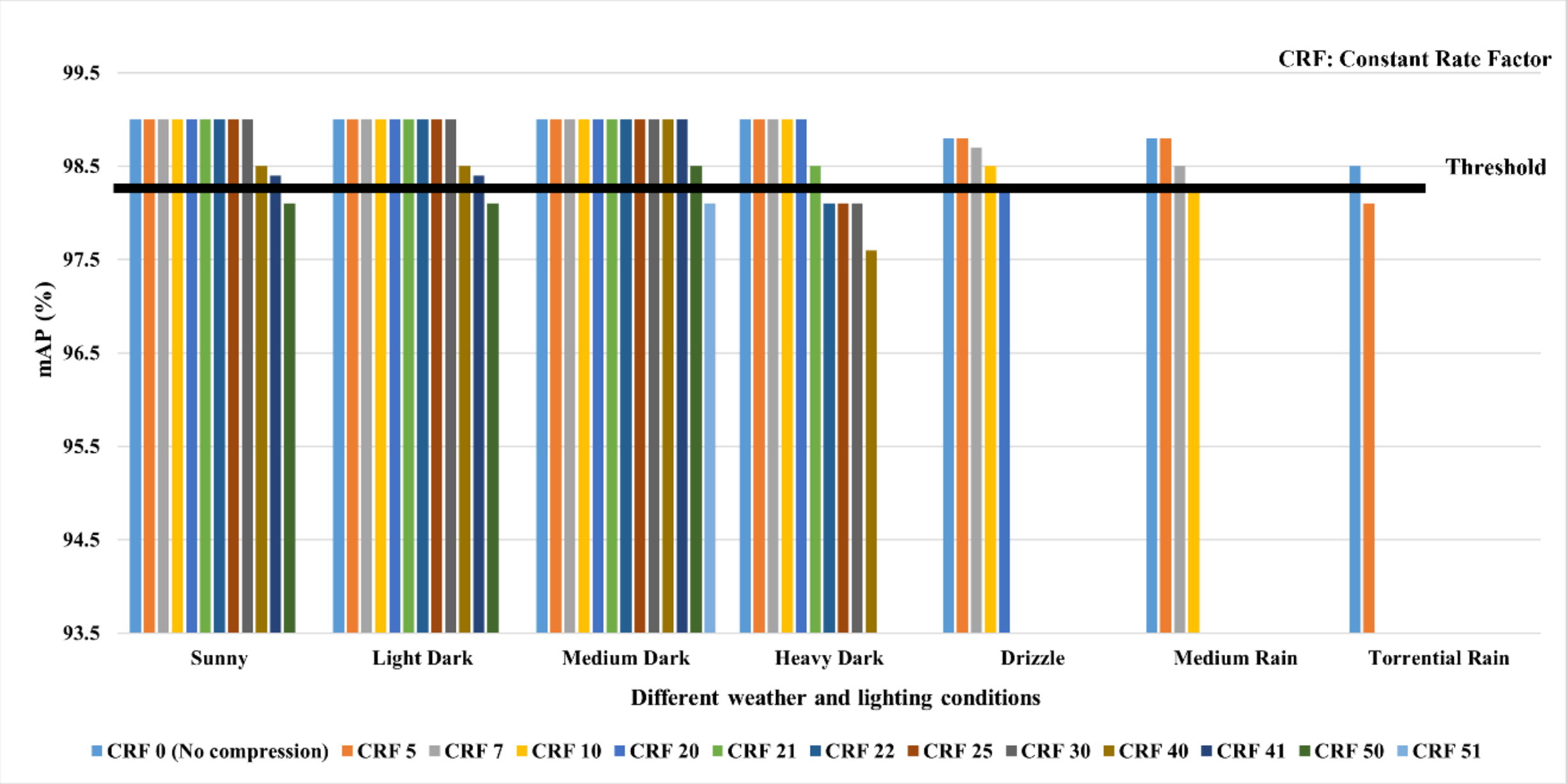

**Figure 6 Vehicle detection mAP for different weather and lighting conditions with different compression levels.**

**Evaluation of PAVC for Bandwidth Reduction**

To evaluate the performance of PAVC on video-based vehicle detection, we conduct field tests under communication facilities with different bandwidths. We utilize two end nodes: one serves as the sending node, responsible for detecting weather conditions and transmitting either compressed or uncompressed videos based on prevailing weather and lighting conditions via Wi-Fi. The other node acts as the receiving node, tasked with receiving the videos and performing vehicle detection to generate Basic Safety Messages using the algorithm outlined in (*9*). However, the PAVC's applicability extends to any vehicle detection-based applications with any communication facilities.

The tests are conducted in three Wi-Fi networks with available communication bandwidths of 5.1 Mbps (decent communication facility), 0.51 Mbps (low communication facility), and 0.065 Mbps (very low communication facility). We follow the strategy as mentioned in **Figure 2** for testing. During video data transmission and reception, we measure the data transmission time for both compressed and uncompressed videos under seven different weather and lighting conditions. **Equation 1** and **Equation 2** represent the data transmission times for the original video and the PAVC method, respectively.

$$T_{trans,org} = \frac{S_{org}}{B_{comm}} \quad \textbf{(1)}$$

$$T_{trans,PAVC} = \frac{S_{org}}{B_{comp}} + \frac{S_{comp}}{B_{comm}} + t_{weatherdetector} \quad \textbf{(2)}$$

here, $T_{trans,org}$ and $T_{trans,PAVC}$ are the video data transmission times for without compression (PAVC) and under PAVC scenarios, respectively; $S_{org}$, $S_{comp}$, $B_{comm}$, $B_{comp}$, and $T_{weatherdetector}$ are original data size, compressed data size, communication bandwidth, compression bandwidth, and weather and lighting conditions detection time, respectively.

File sizes are measured in bytes, while communication is measured in megabits per second (Mbps) and compression bandwidth is measured in megabytes per second (MBps). Using the above two equations, the speed-up for the PAVC method is calculated with the following equation, where a speed-up value greater than 1 indicates that the PAVC method is faster than the regular uncompressed video transmission process.

$$\boldsymbol{Speed_{up}} = \frac{\boldsymbol{T_{trans,org}}}{\boldsymbol{T_{trans,PAVC}}} \quad \textbf{(3)}$$

**Figure 7** illustrates the conditions under which our PAVC proves effective. It demonstrates that PAVC is highly effective under sunny, light-dark, and medium-dark conditions and somewhat effective under heavy-dark weather, particularly when communication bandwidth is severely limited.

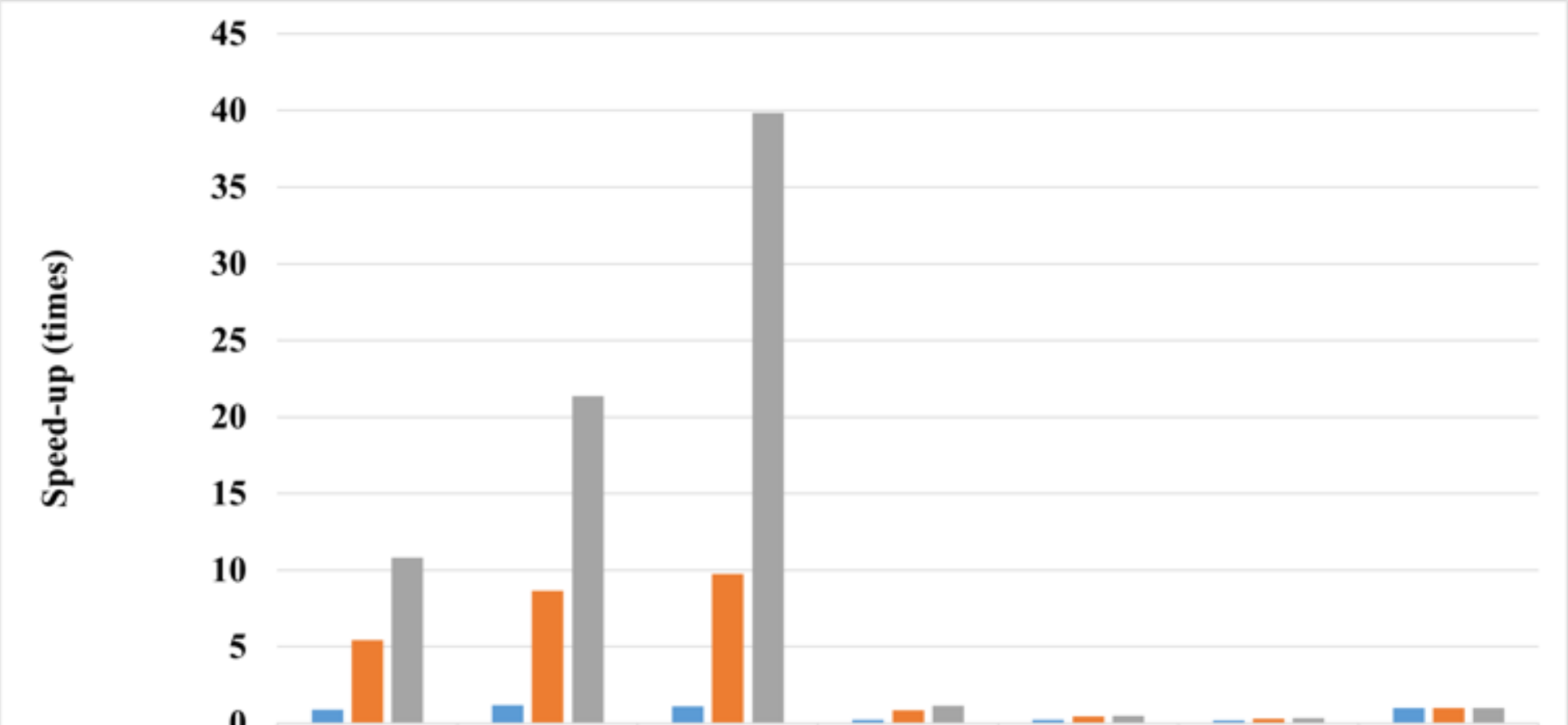

| | Sunny | Light Dark | Medium Dark | Heavy Dark | Drizzle | Moderate Rain | Torrential Rain |
|---|---|---|---|---|---|---|---|
| 5.1 MBPS | 0.884028 | 1.209801 | 1.110875 | 0.248738 | 0.237411 | 0.201885 | 1 |
| 0.51 MBPS | 5.429403 | 8.643321 | 9.761404 | 0.872303 | 0.454282 | 0.330399 | 1 |
| 0.065 MBPS | 10.82598 | 21.3785 | 39.83627 | 1.152386 | 0.498423 | 0.352131 | 1 |

**Figure 7 Data transmission time ratio of uncompressed video data transmission time to PAVC transmission time, demonstrating the speed-up under all weather and lighting conditions.**

However, when the available communication bandwidth is not limited, in our experiment, 5.1 Mbps, we observe no noticeable performance improvement in terms of speed-up, as shown in **Figure 7**. Nevertheless, improvements can still be achieved depending on the compression ratio and compression bandwidth for any given communication bandwidth, including 5.1 Mbps. **Figure 8** illustrates that PAVC can achieve speed-up under any available communication bandwidth. Specifically, **Figure 8** suggests that when communication bandwidth is high, for any given compression ratio, a higher compression speed or bandwidth is required to realize speed-up (as shown in **Figure 8(c)**), compared to when the communication bandwidth is limited (as shown in **Figure 8(a)**). Similarly, when communication bandwidth is high, then for a given compression bandwidth, a higher compression ratio is required to speed up (as shown in **Figure 8(c)**), compared to when the communication bandwidth is limited (as shown in **Figure 8(a)**). Hence, in all cases, speed-ups are achievable in our FAVC.

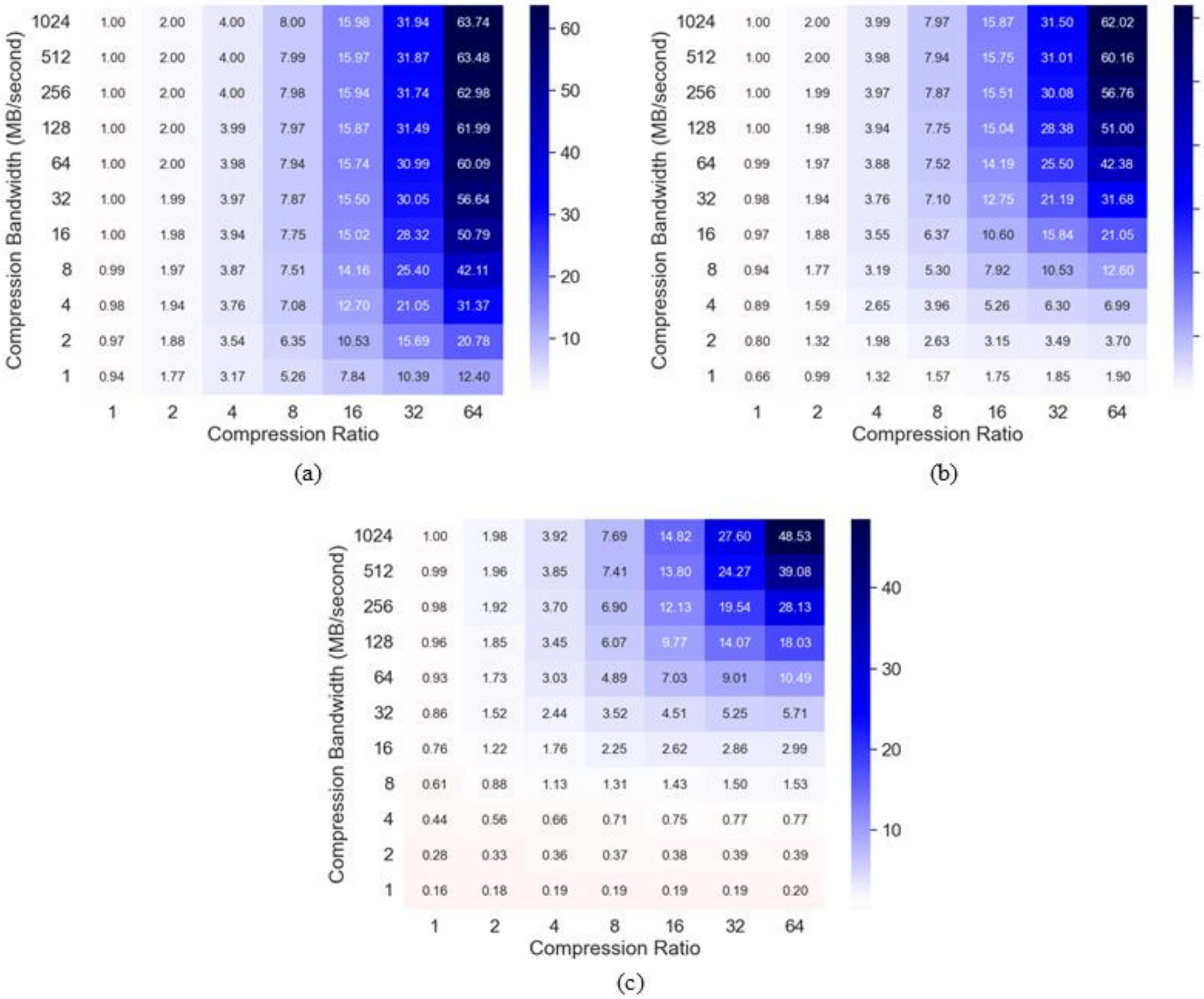

**Figure 8 Speed-up for different compression bandwidth and compression ratio when available communication bandwidth is (a) 0.065 Mbps, (b) 0.51 Mbps, and (c) 5.1 Mbps.**

We further analyze the bandwidth reduction for PAVC under various weather and lighting conditions. For this, we measure the required bandwidth for PAVC when it matches the data transmission times of the uncompressed video transmission. To achieve this, we consider equal data transmission times for both uncompressed and compressed scenarios as per **Equation 1** and **Equation 2**. The results, depicted in **Figure 9,** show that a bandwidth reduction ratio greater than 1 indicates the effectiveness of PAVC. Our findings reveal that PAVC is highly effective for sunny, light-dark, and medium-dark conditions. The results also indicate that PAVC's effectiveness increases significantly when the available communication bandwidth is very low, suggesting that PAVC becomes more beneficial under heavy traffic conditions and/or when high-speed communication is not feasible.

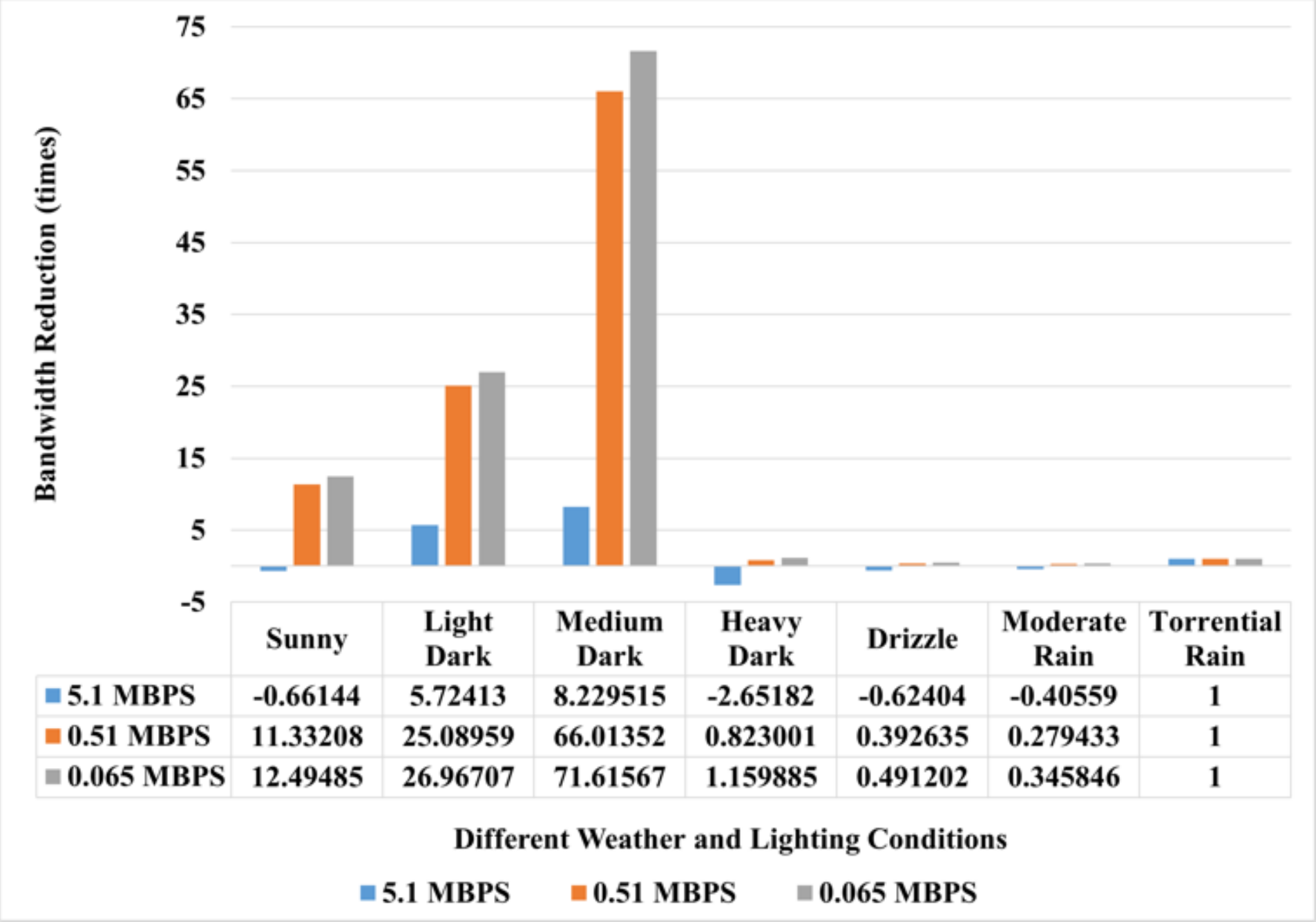

| | Sunny | Light Dark | Medium Dark | Heavy Dark | Drizzle | Moderate Rain | Torrential Rain |
|---|---|---|---|---|---|---|---|
| 5.1 MBPS | -0.66144 | 5.72413 | 8.229515 | -2.65182 | -0.62404 | -0.40559 | 1 |
| 0.51 MBPS | 11.33208 | 25.08959 | 66.01352 | 0.823001 | 0.392635 | 0.279433 | 1 |
| 0.065 MBPS | 12.49485 | 26.96707 | 71.61567 | 1.159885 | 0.491202 | 0.345846 | 1 |

**Figure 9 Bandwidth reduction ratios for PAVC compared to no compression scenarios under all weather and lighting conditions.**

## CONCLUSIONS

This work focuses on developing a precision-aware video compression (PAVC) technique that compresses video to varying levels without compromising object detection accuracy, in this case, mAP in vehicle detection. Object detection performance tends to decrease in poor weather conditions and with reduced video quality. Therefore, it is not feasible to apply the same level of compression to videos under all weather and lighting conditions. Our PAVC strategy dynamically adjusts video quality to ensure that vehicle detection performance does not fall below the threshold mAP of 98.5%. This is achieved by retraining the vehicle detection model, the YOLOv7, using compressed data at different compression levels under various weather and lighting conditions. Using this dynamic approach, PAVC reduces video communication bandwidth by up to 72× in areas with limited communication resources while maintaining a mAP of 98.5% for vehicle detection. This method can be applied to any video-based object detection application.

In this study, the image data for diverse weather and lighting conditions is generated through image augmentations from solely sunny weather and lighting conditions data. As a result, images under a particular environmental condition exhibit uniformity, which might not be the case in real-world scenarios. Future work could focus on collecting image data for all weather and lighting conditions. Based on the collected data, different environmental conditions could be categorized. Then, the PAVC method could be calibrated, and the compression levels for different environmental conditions could be evaluated and adjusted.

## ACKNOWLEDGMENTS

This work is based upon the work supported by the National Center for Transportation Cybersecurity and Resiliency (TraCR) (a U.S. Department of Transportation National University Transportation Center) headquartered at Clemson University, Clemson, South Carolina, U.S., and the U.S. National Science Foundation (NSF). Any opinions, findings, conclusions, and recommendations expressed in this material are those of the author(s) and do not necessarily reflect the views of TraCR, NSF, and the U.S. Government assumes no liability for the contents or use thereof.

Note that ChatGPT was used solely to check grammar and paraphrase texts. No information, text, figure, or table has been generated, nor has any kind of analysis been conducted using any Large Language Model or Generative Artificial Intelligence.

## AUTHOR CONTRIBUTIONS

The authors confirm contribution to the paper as follows: study conception and design: A. Enan, J. Calhoun, M. Chowdhury; data collection: A. Enan; analysis: A. Enan; interpretation of results: A. Enan, J. Calhoun, M. Chowdhury; draft manuscript preparation: A. Enan, J. Calhoun, M. Chowdhury. All authors reviewed the results and approved the final version of the manuscript.

## DECLARATION OF CONFLICTING INTERESTS

The authors declared no potential conflicts of interest with respect to the research, authorship, and/or publication of this article.

## FUNDING

This research was supported by the National Center for Transportation Cybersecurity and Resiliency (TraCR) (grant no. 69A3552344812 and 69A3552348317), and the National Science Foundation (NSF) (grant no. SHF-1943114 and SHF-2312616).